\documentclass[letterpaper, 10 pt, conference]{ieeeconf}  

\IEEEoverridecommandlockouts                              

\IfFileExists{xcolor.sty}{%
\usepackage[table, dvipsnames]{xcolor}
}{%
\usepackage{color}
\definecolor{Red}{rgb}{1,0,0}
\definecolor{orange}{rgb}{1,0.5,0}
\definecolor{purple}{rgb}{0.5,0,0.5}
\definecolor{magenta}{rgb}{1,0,1}
\definecolor{cyan}{rgb}{0,1,1}
\definecolor{blue}{rgb}{0,0,1}
}
\usepackage{amsmath} 
\usepackage{graphicx}
\IfFileExists{caption.sty}{\usepackage{caption}}{\newcommand{\captionsetup}[1]{}}
\IfFileExists{amssymb.sty}{\usepackage{amssymb}}{}
\IfFileExists{xspace.sty}{\usepackage{xspace}}{\newcommand{\xspace}{}}
\providecommand{\href}[2]{#2}
\providecommand{\url}[1]{#1}
\IfFileExists{cite.sty}{\usepackage{cite}}{}
\IfFileExists{multirow.sty}{\usepackage{multirow}}{\newcommand{\multirow}[3]{##3}}
\IfFileExists{float.sty}{\usepackage{float}}{}

\title{\LARGE \bf
DART: Learning-Enhanced Model Predictive Control for Dual-Arm Non-Prehensile Manipulation
}

\author{Autrio Das, Shreya Bollimuntha, Madala Venkata Renu Jeevesh,\\
Keshab Patra, Tashmoy Ghosh, Nagamanikandan Govindan,\\
Arun Kumar Singh, and K Madhava Krishna}

\newcommand{\acro}{DART\xspace}

\begin{document}

\maketitle
\thispagestyle{empty}
\pagestyle{empty}

\begin{abstract}

\textit{What appears effortless to a human waiter remains a major challenge for robots.}
Manipulating objects non-prehensilely on a tray is inherently difficult, and the complexity is amplified in dual-arm settings. Such tasks are  highly relevant to service robotics in domains such as hotels and hospitality, where robots must transport and reposition diverse objects with precision. We present \acro, a novel dual-arm framework that integrates nonlinear Model Predictive Control (MPC) with an optimization-based impedance controller to achieve accurate object motion relative to a dynamically controlled tray. The framework systematically evaluates three complementary strategies for modeling tray–object dynamics as the state transition function within our MPC formulation: (i) a physics-based analytical model, (ii) an online regression-based identification model that adapts in real-time, and (iii) a reinforcement learning–based dynamics model that generalizes across object properties.
Our pipeline is validated in simulation with objects of varying mass, geometry, and friction coefficients. Extensive evaluations highlight the trade-offs among the three modeling strategies in terms of settling time, steady-state error, control effort, and generalization across objects. To the best of our knowledge, \acro constitutes the first framework for non-prehensile Dual-Arm manipulation of objects on a tray. Project Link: \href{https://dart-icra.github.io/dart/}{https://dart-icra.github.io/dart/}


\end{abstract}

\section{INTRODUCTION}
Service robots performing human like tasks face a fundamental limitation: most rely exclusively on grasping to manipulate objects~\cite{garmi}, struggling with items that are large, fragile, or awkwardly positioned i.e. necessitating non-prehensile manipulation, i.e., manipulating objects without grasping~\cite{hacman, mason1999progress}. This field encompasses diverse strategies including pushing~\cite{pizzapeel}, throwing~\cite{pizza_tossing}, and rolling~\cite{np_rolling_passivity}, exploiting friction, inertia, and contact forces without rigid grasping to offer promising alternatives for challenging scenarios. Among these techniques~\cite{mason1999progress,pizza_tossing,yu2016more,np_rolling_passivity,RL-NP2025RAL,single_arm}, tray-based object manipulation moves objects through controlled surface tilting, presenting a natural, human-inspired approach with broad applications in restaurants, hospitals, and laboratories. However, extending this capability to dual-arm systems introduces various challenges: the shared tray dynamically couples both arms, requiring precise coordination while simultaneously controlling object motion through uncertain dynamics. Unlike single-arm manipulation, this problem demands understanding and predicting how tray tilting affects object trajectories while maintaining stable, compliant control of the two manipulators.

When two robotic arms coordinate to manipulate a tray for controlling object motion on its surface, several critical constraints emerge~\cite{dual-arm-survey}. \textbf{First}, the tray creates dynamic coupling between the arms, meaning any motion by one arm directly affects the tray orientation and, consequently, the object's trajectory and the other arm's required response. \textbf{Second}, precise object control requires accurate modeling of complex tray–object dynamics involving friction, inertia, and contact mechanics that vary significantly with object properties such as mass, geometry, and surface characteristics. \textbf{Third}, the system must maintain compliance to handle interaction forces, while executing coordinated tilting motions, as rigid control can amplify disturbances and destabilize the entire manipulation process. These coupled dynamics create a challenging control problem where traditional approaches struggle with adaptability.
\begin{figure}[t]
    \vspace{7pt}
    \centering
    \captionsetup{font=footnotesize}
    \includegraphics[width=0.9\linewidth]{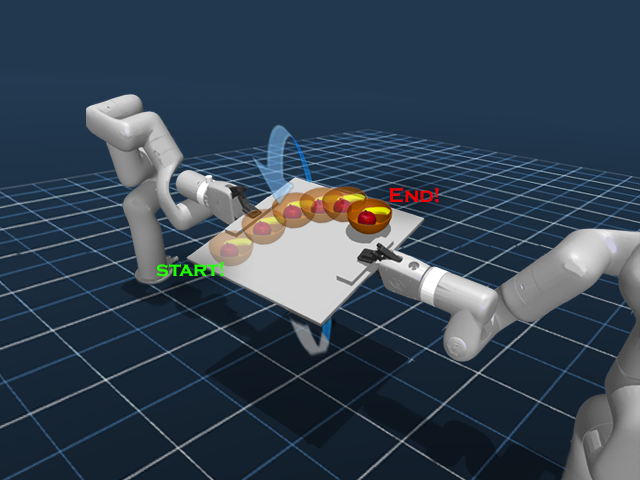}
    \caption{\textbf{Non-Prehensile Manipulation} using a Dual xArm7 setup. Our framework transports a bowl of fruits to the goal location on the tray through a sequence of tilts given by MPC executed by the dual-arm system.}
    \label{fig:enter-label}
\end{figure}

Existing research has largely treated non-prehensile manipulation and dual-arm coordination as separate problems. Non-prehensile manipulation work has primarily focused on single-arm systems~\cite{pizzapeel,keep_it_upright,selvaggio2023non}, while dual-arm coordination research typically assumes grasped objects with well-defined contact models~\cite{dual_arm_benchmark,bimanual-taxonomy,da2,dg16m}. The intersection of dual-arm coordination and non-prehensile manipulation remains largely unexplored, despite its practical importance and technical complexity. Accurately modeling object-tray dynamics presents a fundamental challenge: analytical models may oversimplify complex contact mechanics, while purely data-driven approaches may lack interpretability and fail to generalize across diverse objects and conditions.

To address these limitations, we propose \acro{} (Dual-Arm Robot Tray manipulation), a unified framework that integrates nonlinear Model Predictive Control (MPC) with optimization-based impedance control. Our MPC computes optimal tray tilt trajectories to guide objects to desired goals while accounting for system constraints and object dynamics. Simultaneously, our impedance controller ensures stable, compliant execution by both arms, managing interaction forces and maintaining coordination throughout the manipulation process. 
To summarize, our contributions are:
\begin{enumerate}
    \item We introduce the to the best of our knowledge a first framework for dual-arm non-prehensile tray manipulation, formulating a control problem that bridges bimanual coordination with contact-rich object manipulation.
    \item We develop three complementary dynamics modeling strategies: physics-based, online regression-based, and reinforcement learning-based. A key innovation of our framework is the integration of RL-learned state transition dynamics as a constraint within the MPC, enabling the controller to generalize across varying object geometries, masses, and surface interactions without requiring elaborate fine-tuning for each new object. 
    
    \item We demonstrate through extensive experiments across objects with varying mass, geometry, and surface properties that our framework performs well. We analyze and compare these approaches across metrics including settling time, steady-state error, and control effort.
\end{enumerate}

\section{RELATED WORKS}
\subsection{Non-Prehensile Manipulation}
Non-prehensile manipulation is crucial for tasks involving hard-to-grasp objects or cluttered workspaces~\cite{hacman, mason1999progress}. This field encompasses diverse techniques such as throwing~\cite{pizza_tossing}, pushing~\cite{pushing_refs}, rolling~\cite{np_rolling_passivity}, and sliding~\cite{RL-NP2025RAL, shi2017dynamic}. Several studies have explored the non-prehensile manipulation of objects on horizontal surfaces~\cite{selvaggio2023non, keep_it_upright, single_arm}. 

Prior research on planar object transportation has followed two distinct approaches. One focuses on maintaining object stability during transport, explicitly preventing sliding or slippage~\cite{selvaggio2023non,keep_it_upright}. In contrast, other works deliberately exploit sliding as the manipulation mechanism~\cite{RL-NP2025RAL}. Many previous studies assume the presence of external barriers to constrain object motion and rely on precise knowledge of surface friction coefficients within their model-based controllers~\cite{shi2017dynamic}. While such setups enable impressive controllability, they limit generalization to unconstrained environments with unknown friction properties. Recent advances have addressed these limitations through adaptive and learning-based approaches~\cite{RL-NP2025RAL}, but the vast majority of non-prehensile manipulation research remains constrained to single-arm systems.

\subsection{Dual-Arm Manipulation}
Dual-arm manipulation extends robotic capabilities beyond single-arm limitations, enabling coordinated object transport and bimanual assembly~\cite{dual-arm-survey,bimanual-taxonomy}. Dual-arm systems are pivotal for manipulating heavier payloads that exceed single-arm capacity limits~\cite{two_arms_better,dual_vs_single}, providing enhanced stability through distributed load-bearing, crucial for tray based transportation and manipulation where payload weight, tray stability, and precise orientation control must be simultaneously managed~\cite{dual-arm-survey}.

Traditional approaches utilize impedance-based control~\cite{tran_mecha_opti,bimanual_imp_2013} to handle dynamic coupling between manipulators, while recent advances have employed learning methods including imitation learning~\cite{aloha}, vision-language models~\cite{bi_vla}, and reinforcement learning~\cite{bidex}. These approaches have demonstrated impressive capabilities in coordinated manipulation tasks requiring precise force regulation and synchronization between arms~\cite{davil,dualarm_slabstone}, which is essential for maintaining stable tray orientation during transport.

However, existing dual-arm research exclusively assumes rigidly grasped objects with controlled dynamics. This fundamental assumption fails in tray-based scenarios where objects slide freely, creating entirely different contact dynamics that existing frameworks do not address. Single-arm non-prehensile approaches cannot leverage the enhanced force distribution and coordination that dual-arm systems provide.

To bridge this gap, our \acro{} framework introduces the first integration of non-prehensile tray manipulation with dual-arm coordination, combining complementary dynamics modeling strategies within a unified nonlinear MPC and optimization-based impedance control architecture for dual-arm non-prehensile object manipulation. The framework's generality is a key strength, enabling adaptation to diverse object geometries, masses, and surface properties without requiring extensive recalibration or object-specific modeling. \acro{} enables precise manipulation of objects through controlled tray tilting, a capability essential for service robotics applications in domains such as hospitality, healthcare, and domestic environments.

\section{SYSTEM DESCRIPTION}
\begin{figure}[h]
    \centering
    \captionsetup{font=footnotesize}
    \includegraphics[width=0.85\linewidth]{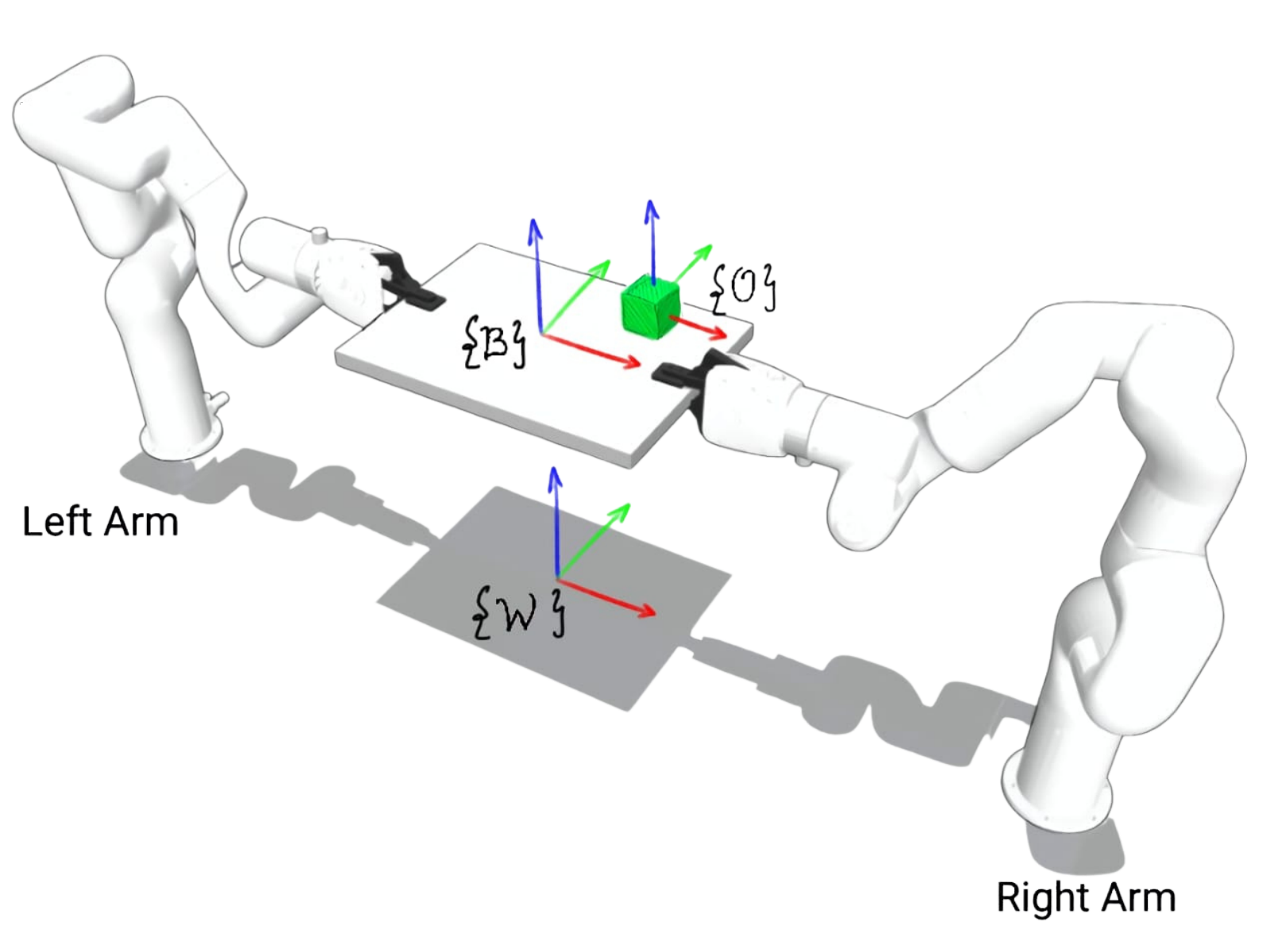}
    \caption{\textbf{Task Setup:} Two robotic arms are placed with the bases fixed to the ground. The tray is rigidly grasped and an object is placed on the tray. The world frame is represented by \(\{\mathcal{W}\}\), body frame by \(\{\mathcal{B}\}\) and object frame by \(\{\mathcal{O}\}\).}
    \label{fig:frames}
\end{figure}

We consider a system of two manipulator arms rigidly grasping a tray as shown in Fig. \ref{fig:frames}, and an object of mass \(m\) is placed on it. The world frame \(\{\mathcal{W}\}\) is fixed to the environment, the tray frame \(\{\mathcal{B}\}\) aligned with \(\{\mathcal{W}\}\) at zero tilt, and the object frame \(\{\mathcal{O}\}\) is attached to the object’s center of mass (CoM). The object-tray configuration and dual arm system are described as follows.
\setlength{\belowcaptionskip}{-6pt}

\subsection{Object–Tray Model} 

\label{sec:system_model_object_tray}

The pose of the object in \(\{\mathcal{W}\}\) is defined as $\boldsymbol{\mathrm{x}} = [\boldsymbol{\mathrm{p}}^\top,\ \boldsymbol{\mathrm{\theta}}^\top]^\top$, where $\boldsymbol{\mathrm{p}} \in \mathbb{R}^3$ is the position and $\boldsymbol{\mathrm{\theta}} \in \mathbb{R}^3$ is the orientation (roll-pitch-yaw).
The linear and angular velocities are $\boldsymbol{\mathrm{v}} \in \mathbb{R}^3$ and $\boldsymbol{\mathrm{\omega} \in  \mathbb{R}^3}$. The twist in \(\{\mathcal{W}\}\) is defined as $\boldsymbol{\mathrm{\nu}} = [\boldsymbol{\mathrm{v}}^\top, \boldsymbol{\mathrm{\omega}}^\top]^\top$. The object state in \(\{\mathcal{W}\}\) is defined as \(\boldsymbol{\mathrm{X}} = [\boldsymbol{\mathrm{x}}^\top, \boldsymbol{\mathrm{\nu}}^\top]^\top \).
The control input \(\boldsymbol{\mathrm{{u}}} = [{\mathrm{{u}_{\alpha}}},\ {\mathrm{{u}_{\beta}}}]^{\top}\) \( \in \mathbb{R}^{2}\) represents the commanded roll (\(\mathrm{u}_{\alpha}\)) and pitch (\(\mathrm{u}_{\beta}\)) angles of the tray in the world frame \(\{\mathcal{W}\}\). 
These tilt angles are constrained by the tray actuation limits \(|\mathrm{u}_{\alpha}|\leq \mathrm{u}_{\alpha}^{\max},\ |\mathrm{u}_{\beta}|\leq \mathrm{u}_{\beta}^{\max}\) (refer to Table~\ref{tab:controller-params}). The desired goal state of the object in \(\mathcal{\{W\}}\) is denoted by $\boldsymbol{\mathrm{X}}^\mathrm{ref}$. The object pose and twist in \(\{\mathcal{O}\}\) are denoted by \(\boldsymbol{\mathrm{x}}_\mathrm{o}\) and \(\boldsymbol{\mathrm{\nu}}_\mathrm{o}\) respectively. Friction coefficient between the object and tray is represented by \(\mu\). 

\subsection{Dual-Robot Model} 
\label{sec:robot-model}
We consider a system of two torque-controlled manipulators with 7 degrees of freedom (DoF).
The joint positions, velocities, and acceleration vectors are denoted by 
\(\boldsymbol{\mathrm{q}}_{\text{L/R}},\ \dot{\boldsymbol{\mathrm{q}}}_{\text{L/R}},\ \ddot{\boldsymbol{\mathrm{q}}}_{\text{L/R}}\),
where the subscript \(\text{L/R}\) refers to the \textit{left} or \textit{right} arm, respectively. 
The joint-space mass matrix is \(\boldsymbol{M}(\boldsymbol{\mathrm{q}})\), and the combined centripetal, Coriolis, and gravitational effects are given by 
\(\boldsymbol{h}(\boldsymbol{\mathrm{q}},\boldsymbol{\mathrm{\dot{q}}})\). The manipulator's Jacobian and its derivative are given by \(\boldsymbol{J}(\boldsymbol{\mathrm{q}})\) and \(\boldsymbol{\dot{J}}(\boldsymbol{\mathrm{q}})\) respectively.
The joint torques applied to each arm are \(\boldsymbol{\mathrm{\tau}}_{\text{L/R}}\).

The two end-effectors rigidly grasp the tray at fixed, predefined grasp poses \(\{G_L,G_R\}\in SE(3)\), enabling coordinated tray manipulation. The poses of the end-effector are expressed as
\(
\boldsymbol{\mathrm{y}}_{\text{L/R}} = 
\begin{bmatrix}
\boldsymbol{\mathrm{y}}_{\text{pos,L/R}}, \quad
\boldsymbol{\mathrm{y}}_{\text{ori,L/R}}
\end{bmatrix}^\top,
\)
where \(\boldsymbol{\mathrm{y}}_{\text{pos,L/R}}\in\mathbb{R}^3\) is the Cartesian position of the end-effector 
and \(\boldsymbol{\mathrm{y}}_{\text{ori,L/R}}\in\mathbb{R}^3\) is its orientation (e.g., roll–pitch–yaw) in \(\{\mathcal{W}\}\).  The desired end–effector poses \(\boldsymbol{\mathrm{y}}^{\text{ref}}_{\text{L/R}}\) are obtained from the desired tray pose as their relative placement w.r.t. the tray CoM stays constant.


\section{METHOD}
\label{sec:method}

We adopt a two-layer control framework for non-prehensile manipulation of the object as described below:
\begin{enumerate}
    \item \textbf{High-level controller:} 
    A nonlinear MPC (refer Sec.~\ref{sec:mpc}) for manipulating the object from its current state $\boldsymbol{\mathrm{X}}$ to the desired state $\boldsymbol{\mathrm{X}}^{\mathrm{ref}}$, computes \(\boldsymbol{\mathrm{u}}\) over the prediction horizon $\mathrm{N}$. 
    The dynamics rollout constraint in this MPC is modeled in three different ways as described in Sec.~\ref{sec:object-tray-dynamics}. 
    
    \item \textbf{Low-level controller:} 
    A torque controller, adapted from an optimization-based dual-arm impedance control framework (refer to Sec.~\ref{sec:davil-controller})~\cite{QP_for_dualarm,davil}, computes \(\boldsymbol{\mathrm{\tau}}_{\text{L/R}}\) for each manipulator in order to realize the commanded tray tilts \(\boldsymbol{\mathrm{u}}\), as given by the high-level controller. 
\end{enumerate}

\subsection{Model Predictive Control}
\label{sec:mpc}

The high-level MPC controller computes optimal control commands $\boldsymbol{\mathrm{u}}_{\mathrm{k}}$ and  object states $\boldsymbol{\mathrm{X}}_{\mathrm{k}}$ for a horizon of length $\mathrm{N}$. It solves the nonlinear optimization in Eq.~\ref{eqn:MPC} for each discrete time step $\mathrm{k}\in[0,\mathrm{N})$ which is sampled at \(\mathrm{T}_{\mathrm{s}}\).
The tracking error at step \(\mathrm{k}\) is defined as \(\boldsymbol{\mathrm{e}}_{\mathrm{k}}=(\boldsymbol{\mathrm{X}}_{\mathrm{k}}-\boldsymbol{\mathrm{X}}^{\mathrm{ref}})\).


 The cost function penalizes the state tracking error \(\boldsymbol{\mathrm{e}}_{\mathrm{k}}\), the control effort \(\boldsymbol{\mathrm{u}}_{\mathrm{k}}\), the control rate change \(\Delta\boldsymbol{\mathrm{u}}_{\mathrm{k}}\), and the terminal state error \(\boldsymbol{\mathrm{e}}_{\mathrm{N}}\) to ensure smooth tray motion:
\begin{subequations}
\begin{align}
\min_{\mathcal{X},\ \mathcal{U}}\ 
\sum_{\mathrm{k=0}}^{\mathrm{N-1}} & \left( \boldsymbol{\mathrm{e}}_\mathrm{k}^\top \boldsymbol{\mathrm{Q}}\,\boldsymbol{\mathrm{e}}_\mathrm{k} + 
\begin{bmatrix}
    \boldsymbol{\mathrm{u}}_\mathrm{k} \\ \Delta\boldsymbol{\mathrm{u}}_\mathrm{k}  
\end{bmatrix}^{\top}\! \boldsymbol{\mathrm{Q_R}}
\begin{bmatrix}
    \boldsymbol{\mathrm{u}}_\mathrm{k} \\ \Delta\boldsymbol{\mathrm{u}}_\mathrm{k}       
\end{bmatrix}\right)
+ \boldsymbol{\mathrm{e}}_\mathrm{N}^\top\boldsymbol{\mathrm{Q}}_{\mathrm{N}}\boldsymbol{\mathrm{e}}_\mathrm{N} \label{mpcopt}\\
\textrm{s.t.}\ & \boldsymbol{\mathbf{X}}_{\mathrm{k+1} }= \Phi\!\bigl(\boldsymbol{\mathrm{X}}_{\mathrm{k}},\boldsymbol{\mathrm{u}}_{\mathrm{k}},\Delta t\bigr),
\label{eq:general-integrator} \\
& \boldsymbol{\mathrm{\nu}}_{{\min}}\le \boldsymbol{\mathrm{{\nu}}}_{\mathrm{k}} \le \boldsymbol{\mathrm{\nu }}_{\max}
\label{mpccon1}\\
 & \boldsymbol{\mathrm{u}}_{{\min}}\le \boldsymbol{\mathrm{{u}}}_{\mathrm{k}} \le \boldsymbol{\mathrm{u }}_{\max} \label{mpccon3}\\ 
& \Delta \boldsymbol{\mathrm{u}}_{\min}\le \Delta \boldsymbol{\mathrm{{u}}}_\mathrm{k} \le \Delta \boldsymbol{\mathrm{u}}_{\max}\label{mpccon4}
\end{align} \label{eqn:MPC}
\end{subequations}
Here, \(\boldsymbol{\mathrm{Q}}\) and \(\boldsymbol{\mathrm{Q}}_{\mathrm{N}}\) denote the weight matrices for the state tracking error along the horizon and at the terminal step, respectively. In our implementation, \(\boldsymbol{\mathrm{Q}}\) is structured to separately weight position and velocity errors (\(\boldsymbol{\mathrm{Q}}=\mathrm{diag}(\boldsymbol{\mathrm{Q}}_{\mathrm{p}},\boldsymbol{\mathrm{Q}}_{\mathrm{v}})\)) (refer to Table~\ref{tab:controller-params}). The matrix \(\boldsymbol{\mathrm{Q_R}}\) penalizes both the control input \(\boldsymbol{\mathrm{u}}_{\mathrm{k}}\) and its rate of change \(\Delta\boldsymbol{\mathrm{u}}_{\mathrm{k}}\), where 
\begin{equation}   
\Delta \bf u_\mathrm{k} =
\begin{aligned}[c]
\begin{cases}
\mathbf {u_0 - u_{\text{prev}}} , & \mathrm{k}=0,\\
\mathbf u_\mathrm{k} - \mathbf u_{\mathrm{k-1}}, & 0<\mathrm{k}\leq \mathrm{N}
\end{cases}
\end{aligned}
\end{equation}
Here \(\mathbf{u_{\text{prev}}}\) is defined as the command executed at the previous simulation timestep. Initially, \(\mathbf{u_{\text{prev}}}\) is defined as \([0,0]^\top\).
The decision variables are stacked as  
\(\mathcal{X}=[\boldsymbol{\mathrm{{X}}}_{0}^\top,\ \boldsymbol{\mathrm{{X}}}_{1}^\top,\ \dots,\ \boldsymbol{\mathrm{{X}}}_{\mathrm{N}}^\top]^\top\) and  
\(\mathcal{U}=[\boldsymbol{\mathrm{{u}}}_0^\top,\ \boldsymbol{\mathrm{{u}}}_1^\top,\ \dots,\ \boldsymbol{\mathrm{{u}}}_{\mathrm{N}}^\top]^\top\), which together describe the object state and control trajectories over the horizon. The object dynamics are enforced as a discrete-time state-transition mapping \(\Phi(\cdot)\) (Eq.~\ref{eq:general-integrator}). We obtain this function by numerical integration of the object dynamic models illustrated in Sec.~\ref{sec:object-tray-dynamics}.

\begin{figure*}[ht]
    \centering
    \captionsetup{font=footnotesize}
    \includegraphics[width=0.85\linewidth]{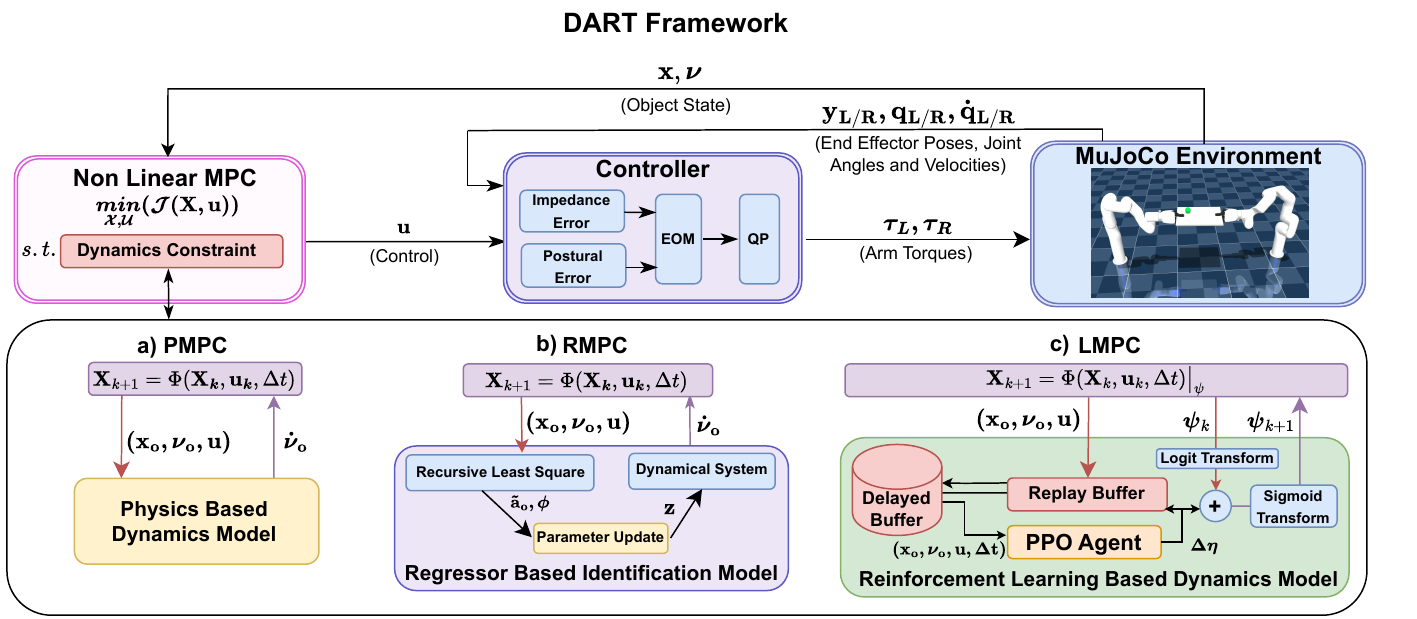}
    \caption{\textbf{DART Framework:} Our proposed framework takes the current object state 
    $\boldsymbol{\mathrm{X}}$
    and the desired target state $\boldsymbol{\mathrm{X}^\text{ref}}$. We choose $\boldsymbol{\nu}^\mathrm{ref}$ as $\boldsymbol{0}_{6\times1}$
    as inputs. These are fed into a nonlinear MPC, which computes the optimal tray-tilt commands (\(\boldsymbol{\mathrm{u}}\)). These commands are then passed to an optimization-based impedance controller, which computes the torques required to realize the tilts. Feedback from the simulator updates the object state for closing the loop for the next MPC step. The object-tray dynamics is modeled as a state transition constraint for the MPC. We propose three models for this state transition constraint, namely \textbf{(a) PMPC:} an analytical physics-based dynamics model, \textbf{(b) RMPC:} a regressor-based model which learns unmodeled dynamics and \textbf{(c) LMPC:} a PPO agent used to estimate the object dynamics.
    }
    \label{fig:pipeline}
\end{figure*}


\subsection{Object–Tray Dynamics}
\label{sec:object-tray-dynamics}

Within our MPC framework (Sec.~\ref{sec:mpc}), the state–transition function (Eq.~\ref{eq:general-integrator}) is instantiated with three alternative dynamics models.  We denote the complete MPC pipeline under each instantiation as :

\begin{itemize}
    \item \textbf{Physics–Based MPC (PMPC):} Uses analytical object dynamics with simple contact models and therefore requires modeling the dynamics of the object. 
    
    \item \textbf{Regression–Based MPC (RMPC):} Augments a nominal dynamics model with an online linear regressor to capture unmodeled dynamics and contact effects, reducing reliance on exact dynamics.
    
    \item \textbf{Reinforcement–Learning–Based MPC (LMPC):} Starts from a nominal dynamics model and learns object dynamics parameters via reinforcement learning (RL) to directly capture unmodeled behaviors.
\end{itemize}

PMPC is purely physics–based, whereas RMPC and LMPC relax the need for exact dynamics and contact models through online adaptation or learning. The three variants are detailed below.

\subsubsection{\textbf{Physics-Based Dynamics Model}}

The object dynamics in the frame $\{\mathcal{O}\}$ as adapted from~\cite{subburaman2023non} is expressed as
\begin{equation}
    \boldsymbol{\mathrm{M}}_\mathrm{o}(\boldsymbol{\mathrm{x}}_\mathrm{o})\,
    \dot{\boldsymbol{\nu}}_\mathrm{o} +
    \boldsymbol{\mathrm{C}}_\mathrm{o}(\boldsymbol{\mathrm{x}}_\mathrm{o}, \boldsymbol{\nu}_\mathrm{o})\, 
    \boldsymbol{\nu}_\mathrm{o} +
    \boldsymbol{\mathrm{g}}_\mathrm{o}(\boldsymbol{\mathrm{x}})
    = \boldsymbol{\mathrm{F}}_\mathrm{o}(\boldsymbol{\mathrm{x}}_\mathrm{o},\boldsymbol{\nu}_\mathrm{o},\boldsymbol{\mathrm{u}})
    \label{eq:EOM}
\end{equation}
where 
$\boldsymbol{\mathrm{M}}_\mathrm{o} \in \mathbb{R}^{6 \times 6}$, 
$\boldsymbol{\mathrm{C}}_\mathrm{o} \in \mathbb{R}^{6 \times 6}$, and 
$\boldsymbol{\mathrm{g}}_\mathrm{o} \in \mathbb{R}^{6}$ denote, respectively, the inertia matrix, the centrifugal and Coriolis matrix, and the gravity vector of the object. Here $\boldsymbol{\mathrm{F}}_\mathrm{o}(\boldsymbol{\mathrm{x}}_\mathrm{o},\boldsymbol{\nu}_\mathrm{o},\boldsymbol{\mathrm{u}}) = \boldsymbol{\mathrm{J}}_\mathrm{c}(\boldsymbol{\mathrm{x}}_\mathrm{o})^\top \boldsymbol{f}(\boldsymbol{\mathrm{x}}_\mathrm{o},\boldsymbol{\nu}_\mathrm{o},\boldsymbol{\mathrm{u}}) \in \mathbb{R}^6$ denotes the generalized contact wrench produced due to the tray's contact interactions. $\boldsymbol{\mathrm{J}}_\mathrm{c}(\boldsymbol{\mathrm{x}}_\mathrm{o})$ represents the contact Jacobian and $\boldsymbol{f}(.)$ encodes the stacked contact forces. Frictional effects are modeled using the Stribeck friction model~\cite{friction}.

 

The tray orientation (tilt) is controlled through the dual-arm manipulators. The tilt $\boldsymbol{\mathrm{u}}$ modifies the components of the gravity vector acting along the tray’s tangent plane.  
Thus, $\boldsymbol{\mathrm{g}}_\mathrm{o}(\boldsymbol{\mathrm{x}}_\mathrm{o})$ depends on the tray tilt and we can say
$\boldsymbol{\mathrm{g}}_\mathrm{o}(\boldsymbol{\mathrm{x}}_\mathrm{o}) \sim \boldsymbol{\mathrm{G}}_\mathrm{o} (\boldsymbol{\mathrm{x}}_\mathrm{o},\boldsymbol{\mathrm{u}})$. 
After updating  and rearranging, we compute the acceleration of the object as follows:
\begin{gather}
\label{eq:dynamics_eqn}
    \dot{\boldsymbol{\nu}}_\mathrm{o} =
    \boldsymbol{\mathrm{M}}_\mathrm{o}^{-1}(\boldsymbol{\mathrm{x}}_\mathrm{o})
    \Bigl[
        \boldsymbol{\mathrm{F}}_\mathrm{o}(\boldsymbol{\mathrm{x}}_\mathrm{o},\boldsymbol{\nu}_\mathrm{o},\boldsymbol{\mathrm{u}}) -
        \boldsymbol{\mathrm{C}}_\mathrm{o}(\boldsymbol{\mathrm{x}}_\mathrm{o}, \boldsymbol{\nu}_\mathrm{o})\,\boldsymbol{\nu}_\mathrm{o} -
        \boldsymbol{\mathrm{G}}_\mathrm{o}(\boldsymbol{\mathrm{x}}_\mathrm{o},\boldsymbol{\mathrm{u}})
    \Bigr]
\end{gather}
which can be compactly written in continuous time as
\begin{equation}
   \dot{\boldsymbol{\mathrm{\nu}}_\mathrm{o}} = 
   \mathcal{F}\bigl(\boldsymbol{\mathrm{x}}_\mathrm{o},\,\boldsymbol{\mathrm{\nu}}_\mathrm{o},\,\boldsymbol{\mathrm{u}}\bigr)
   \label{eq:object_dynamics}
\end{equation}
Numerical Integration of Eq.~\ref{eq:object_dynamics} yields the state-transition function
\(\Phi(\cdot)\) used as the rollout constraint in the MPC:
\begin{equation}
\label{eq:state_transition}
    \boldsymbol{\mathrm{X}}_{\mathrm{k+1}} =
    \Phi\!\bigl(
        \boldsymbol{\mathrm{X}}_{\mathrm{k}},\,
        \boldsymbol{\mathrm{u}}_{\mathrm{k}}
    \bigr)
\end{equation}
$\Phi$ is computed by transforming the pose from $\{\mathcal{W}\}$ to $\{\mathcal{O}\}$,  evaluating object dynamics in $\{\mathcal{O}\}$, integrating $\boldsymbol{\nu}$ in $\{\mathcal{O}\}$ and then updating the pose in $\{\mathcal{W}\}$ using the appropriate map.\\
\subsubsection{\textbf{Regression-Based Identification Model}}
We employ a regressor-based model, formulated as a Recursive Least Squares (RLS)~\cite{lequy2023stochastic} estimator, to learn the unmodeled dynamics.



For each Cartesian axis
$\mathrm{j}\in\{\mathrm{x},\mathrm{y},\mathrm{z}\}$ in $\{\mathcal{O}\}$,
we construct 3$\times$1 feature vectors $^\mathrm{j}\boldsymbol{\phi}_\mathrm{t} = [\mathrm{v}_\mathrm{j}, \tanh\!\big({\mathrm{v}_\mathrm{j}}/\mathrm{\epsilon}_\mathrm{t}\big), 1]^\top$  and $^\mathrm{j}\boldsymbol{\phi}_\mathrm{r} = [\mathrm{\omega}_\mathrm{j}, \tanh\!\big(\mathrm{\omega}_\mathrm{j}/\mathrm{\epsilon}_\mathrm{r}\big), 1]^\top$ for the translational and rotational components of the unmodeled dynamics respectively. Here, $\mathrm{\epsilon}_\mathrm{t}$ and $\mathrm{\epsilon}_\mathrm{r}$ are regularization terms. \

These vectors ($^\mathrm{j}\boldsymbol{\phi}_\mathrm{t},\ ^\mathrm{j}\boldsymbol{\phi}_\mathrm{r}$) capture Coriolis and damping effects, Coulomb friction and other bias, which correlate with velocity, tanh expressions and a numerical constant respectively.

The adaptive parameter vectors,
$^\mathrm{j}\boldsymbol{\mathrm{z}}_\mathrm{t} ,^\mathrm{j}\boldsymbol{\mathrm{z}}_\mathrm{r} \in \mathbb{R}^3$,
and the feature vectors $^\mathrm{j}\boldsymbol{\phi}_\mathrm{t}\ \ \text{and}\ \ ^\mathrm{j}\boldsymbol{\phi}_\mathrm{r}$
are used to model the dynamics as follows:
\begin{gather}
    \hat{\boldsymbol{\mathrm{a}}}_\mathrm{o} = \overline{\boldsymbol{\mathrm{a}}}_\mathrm{o} +\Delta\hat{\boldsymbol{\mathrm{a}}}_{\mathrm{o},\mathrm{k}} \\
    \overline{\boldsymbol{\mathrm{a}}}_\mathrm{o} =     \boldsymbol{\mathrm{M}}_\mathrm{o}^{-1}(\boldsymbol{\mathrm{x}}_\mathrm{o})
    \Bigl[-
        \boldsymbol{\mathrm{C}}_\mathrm{o}(\boldsymbol{\mathrm{x}}_\mathrm{o}, \boldsymbol{\nu}_\mathrm{o})\,\boldsymbol{\nu}_\mathrm{o} -
        \boldsymbol{\mathrm{G}}_\mathrm{o}(\boldsymbol{\mathrm{x}}_\mathrm{o},\boldsymbol{\mathrm{u}})
    \Bigr]
    \label{en:acc_nominal}
\end{gather}
where
$\Delta\hat{\boldsymbol{\mathrm{a}}}_{\mathrm{o},\mathrm{k}} =$ 
\begin{gather}
\bigl[
^\mathrm{x}\boldsymbol{\phi}^\top_\mathrm{t} {}^\mathrm{x}\boldsymbol{\mathrm{z}}_\mathrm{t}, 
^\mathrm{y}\boldsymbol{\phi}^\top_\mathrm{t} {}^\mathrm{y}\boldsymbol{\mathrm{z}}_\mathrm{t}, 
^\mathrm{z}\boldsymbol{\phi}^\top_\mathrm{t} {}^\mathrm{z}\boldsymbol{\mathrm{z}}_\mathrm{t}, 
^\mathrm{x}\boldsymbol{\phi}^\top_\mathrm{r} {}^\mathrm{x}\boldsymbol{\mathrm{z}}_\mathrm{r}, 
^\mathrm{y}\boldsymbol{\phi}^\top_\mathrm{r} {}^\mathrm{y}\boldsymbol{\mathrm{z}}_\mathrm{r},
^\mathrm{z}\boldsymbol{\phi}^\top_\mathrm{r} {}^\mathrm{z}\boldsymbol{\mathrm{z}}_\mathrm{r}
\bigr]^\top
\end{gather}

\subsubsection*{Online Parameter Identification}
At each step \(\mathrm{k}\), we compute the acceleration discrepancy defined  as
\(
\tilde{\boldsymbol{\mathrm{a}}}_\mathrm{o} =
\begin{bmatrix}\tilde{\boldsymbol{\mathrm{a}}_{\mathrm{o},\mathrm{t}}}^\top,
\tilde{\boldsymbol{\mathrm{a}}_{\mathrm{o},\mathrm{r}}}^\top\end{bmatrix}^\top \in \mathbb{R}^6,\ 
\tilde{\boldsymbol{\mathrm{a}}}_{\mathrm{o},\mathrm{k}} =
\boldsymbol{\mathrm{a}}_{\mathrm{o},\mathrm{k}} -
\overline{\boldsymbol{\mathrm{a}}}_{\mathrm{o},\mathrm{k}}
\)
where $\boldsymbol{\mathrm{a}}_{\mathrm{o},\mathrm{k}} = (\boldsymbol{\nu}_{\mathrm{o},\mathrm{k}} - \boldsymbol{\nu}_{\mathrm{o},\mathrm{k -1}})/\mathrm{T}_{\mathrm{s}}$.

We then update the adaptive parameter vectors
$^\mathrm{j}\boldsymbol{\mathrm{z}}_\mathrm{t}$ and
$^\mathrm{j}\boldsymbol{\mathrm{z}}_\mathrm{r}$ using the regressors
$^\mathrm{j}\boldsymbol{\phi}_{\mathrm{t},\mathrm{k}} =
{}^\mathrm{j}\boldsymbol{\phi}_\mathrm{t}(\boldsymbol{\mathrm{\nu}}_{\mathrm{o},\mathrm{k-1}})$ and
$^\mathrm{j}\boldsymbol{\phi}_{\mathrm{r},\mathrm{k}} =
{}^\mathrm{j}\boldsymbol{\phi}_\mathrm{r}(\boldsymbol{\mathrm{\nu}}_{\mathrm{o},\mathrm{k-1}})$.

For all $\mathrm{j}\in\{\mathrm{x},\mathrm{y},\mathrm{z}\}$ and
forgetting factor $\mathrm{\lambda}\in(0,1]$, the RLS update is: 
\begin{gather}
\notag
^\mathrm{j}\mathrm{K}_{\mathrm{t},\mathrm{k}} = \frac{^\mathrm{j}\mathrm{P}_{\mathrm{t},\mathrm{k-1}}{}^\mathrm{j}\phi_{\mathrm{t},\mathrm{k}}}{\mathrm{\lambda} + {}^\mathrm{j}\phi_{\mathrm{t},\mathrm{k}}^\top {}^\mathrm{j}\mathrm{P}_{\mathrm{t},\mathrm{k-1}}{}^\mathrm{j}\phi_{\mathrm{t},\mathrm{k}}}\\ \notag
    ^\mathrm{j}\mathrm{z}_{\mathrm{t},\mathrm{k}} = {}^\mathrm{j}\mathrm{z}_{\mathrm{t},\mathrm{k-1}} + {}^\mathrm{j}\mathrm{K}_{\mathrm{t},\mathrm{k}}\!\left({}^\mathrm{j}\tilde{{\boldsymbol{\mathrm{a}}}}_{\mathrm{o},\mathrm{t},\mathrm{k}} - {}^\mathrm{j}\phi_{\mathrm{t},\mathrm{k}}^\top {}^\mathrm{j}\mathrm{z}_{\mathrm{t},\mathrm{k-1}}\right)\\ 
    {}^\mathrm{j}\mathrm{P}_{\mathrm{t},\mathrm{k}} = \frac{1}{\mathrm{\lambda}}\!\left(^\mathrm{j}\mathrm{P}_{\mathrm{t},\mathrm{k-1}} - {}^\mathrm{j}\mathrm{K}_{\mathrm{t},\mathrm{k}} {}^\mathrm{j}\phi_{\mathrm{t},\mathrm{k}}^\top {}^\mathrm{j}\mathrm{P}_{\mathrm{t},\mathrm{k-1}}\right)
\end{gather}

Here $^\mathrm{j}\mathrm{P}_{\mathrm{t},\mathrm{k}}$ denotes the error covariance matrix and
$^\mathrm{j}\mathrm{K}_{\mathrm{t},\mathrm{k}}$ is the gain vector for axis $\mathrm{j}$.
An analogous update is applied to $^\mathrm{j}\boldsymbol{\mathrm{z}}_{\mathrm{r},\mathrm{k}}$
for all $\mathrm{j}\in\{\mathrm{x},\mathrm{y},\mathrm{z}\}$.
We use $\dot{\boldsymbol{\nu}}_\mathrm{o} = \hat{\boldsymbol{\mathrm{a}}}_{\mathrm{o},\mathrm{k}}$ and numerically integrate it to obtain the state transition constraint similar to Eq.~\ref{eq:state_transition}.

\subsubsection{\textbf{Reinforcement Learning-Based Dynamics Model}}
 We use proximal policy optimization (PPO)~\cite{ppo}, an RL method to learn the parameter vector $\boldsymbol{\psi}$ which collects the state independent components of $\boldsymbol{\mathrm{M}}_\mathrm{o},\ \boldsymbol{\mathrm{C}}_\mathrm{o},\ \boldsymbol{\mathrm{G}}_\mathrm{o}$ and $\boldsymbol{\mathrm{F}}_\mathrm{o}$.



The system dynamics with the learned parameter vector is represented as 
\begin{gather}
\notag
    \dot{\boldsymbol{\nu}}_o =
    \widehat{\boldsymbol{\mathrm{M}}}_o(\boldsymbol{\psi}, \boldsymbol{\mathrm{x}}_o)^{-1}
    \Bigl[
        \widehat{\boldsymbol{\mathrm{F}}}_o(\boldsymbol{\psi}, \boldsymbol{\mathrm{x}}_o, \boldsymbol{\nu}_o) \\ -
        \widehat{\boldsymbol{\mathrm{C}}}_o(\boldsymbol{\psi}, \boldsymbol{\mathrm{x}}_o, \boldsymbol{\nu}_o)\, \boldsymbol{\nu}_o -
        \widehat{\boldsymbol{\mathrm{G}}}_o(\boldsymbol{\psi}, \boldsymbol{\mathrm{x}}_o, \mathrm{u})
    \Bigr]
\end{gather}

PPO is employed to compute the structured updates of the parameter vector $\boldsymbol{\psi}$, enabling the dynamics model to account for unmodeled disturbances, parametric uncertainty, and model mismatch. 

To ensure that $\widehat{\boldsymbol{M}}_\mathrm{o}$ remains positive definite and $\widehat{\boldsymbol{C}}_\mathrm{o}$ and  $\widehat{\boldsymbol{G}}_\mathrm{o}$ are physically plausible, we constrain $\boldsymbol{\psi}$ by means of a tanh squashing function. Further, to prevent saturation, we adopt a Logit–Sigmoid reparameterization to ensure $\boldsymbol{\psi}$ remains bounded, while the learning process is unconstrained. 
 Each parameter is normalized and mapped into logit space.
\begin{equation}
\boldsymbol{r} = {\boldsymbol{\psi}} \ / \psi_{\max} \quad \quad
 \boldsymbol{\eta} = \ln\left(\boldsymbol{r}\  / \ {(1-\boldsymbol{r})}\right)    
\end{equation}
The policy provides $\Delta \boldsymbol{\eta}$ and we map $\boldsymbol{\eta}' = \boldsymbol{\eta} + \Delta \boldsymbol{\eta}$ back through the sigmoid 
\begin{equation}
\boldsymbol{r}' = \sigma(\boldsymbol{\eta}')  \qquad 
\boldsymbol{\psi}' = \psi_{\max} \boldsymbol{r}'  
\end{equation}
This ensures adaptation occurs in an unconstrained space while the realized parameters remain bounded and stable.

Experience from state transitions and MPC commands, along with sampled actions, is stored in a dense replay buffer for policy updates. A secondary sparse buffer collects samples at a lower rate, promoting generalization across variations in geometry, mass, and friction, enabling PPO to adapt robustly to diverse dynamical conditions.

Finally, the policy receives a reward signal \(\boldsymbol{\mathrm{R}}\)
which encourages low position error and low velocity of the object
simultaneously. It also collects the $\mathrm{L1}$ norm of $\Delta\mathbf{u}$ as a penalty.
\begin{align}
\boldsymbol{\mathrm{R}} &= e^{
    \tfrac{-\lVert \boldsymbol{\mathrm{p}} - \boldsymbol{\mathrm{p}}^{\mathrm {ref}} \rVert_2^{2}}{2\sigma_\mathrm{p}^{2}}}
\left(w_{\mathrm {p}}\ 
 +
w_{\mathrm{v}}\ e^{
    \tfrac{ - \lVert \boldsymbol{\mathrm{v}} - \boldsymbol{\mathrm{v}}^{\mathrm{ref}} \rVert_2^{2}}{2\sigma_{\mathrm v}^{2}}}\right) - \lVert \Delta\boldsymbol{\mathrm{u}} \rVert_1
\end{align}
where $w_\mathrm{p}$ and $w_\mathrm{v}$ are scalar weights that are tuned to the desired scaling between the position and velocity minimization rewards. 

\subsection{Controller}
\label{sec:davil-controller}

At each step, the high–level MPC (Sec.~\ref{sec:mpc}) outputs the commanded tray tilts \(\boldsymbol{\mathrm{u}}\) (roll–pitch in \(\{\mathcal{W}\}\)).  
The tray CoM remains fixed while its orientation follows \(\boldsymbol{\mathrm{u}}\).  

We adapt the QP–based dual–arm impedance controller of~\cite{QP_for_dualarm,davil} to realize these poses.
At each time step, the optimal joint accelerations are computed by solving
\begin{gather}
\notag
\left(\boldsymbol{\mathrm{\ddot{q}}}_L^*, \boldsymbol{\mathrm{\ddot{q}}}_R^*\right)
= \arg\min\limits_{\left(\boldsymbol{\mathrm{\ddot{q}}}_L, \boldsymbol{\mathrm{\ddot{q}}}_R\right)}
 w_{\text{imp}} \!\left(\|\boldsymbol{e}_{\text{imp}_{L}}\|^2_2\!+\!\|\boldsymbol{e}_{\text{imp}_{R}}\|^2_2\right) \notag \\
 + w_{\text{pos}}\!\left( \|\boldsymbol{e}_{\text{pos}_L}\|^2_2\!+\!\|\boldsymbol{e}_{\text{pos}_R}\|^2_2 \right)
\label{eq:QP_problem}
\end{gather}
subject to joint position, velocity, and torque limits.

The impedance–task error \(\boldsymbol{e}_{\text{imp}}\) is
\begin{gather}
\boldsymbol{e}_{\text{imp}} = \boldsymbol{\ddot{\boldsymbol{\mathrm{y}}}}- \boldsymbol{\Lambda}^{-1}\!\left[\boldsymbol{D}\!\left(\dot{\boldsymbol{\mathrm{y}}}^{\text{ref}}-\dot{\boldsymbol{\mathrm{y}}}\right)+\boldsymbol{K}\!\left({\boldsymbol{\mathrm{y}}}^{\text{ref}}-{\boldsymbol{\mathrm{y}}}\right)\right]\quad
\\  
\boldsymbol{\Lambda} = \left(\boldsymbol{J}\boldsymbol{M}^{-1}\boldsymbol{J}^\top\right)^{-1},\quad
\boldsymbol{\ddot{\boldsymbol{\mathrm{y}}}}=\boldsymbol{J}\boldsymbol{\mathrm{\ddot{q}}}+\boldsymbol{\dot J}\boldsymbol{\mathrm{\dot q}}
\end{gather}

The postural–task error biases each arm to its grasp configuration:
\begin{gather}
\boldsymbol{e}_{\text{pos}} = \boldsymbol{\mathrm{\ddot{q}}} 
 - 2\sqrt{\boldsymbol{K}_{\text{null}}}\,(\boldsymbol{\mathrm{\dot{q}}}^{\text{ref}}-\boldsymbol{\mathrm{\dot{q}}})
 + \boldsymbol{K}_{\text{null}}(\boldsymbol{\mathrm{q}}^{\text{ref}}-\boldsymbol{\mathrm{q}}),
\end{gather}
with \(\boldsymbol{K},\boldsymbol{K}_{\text{null}}\) and \(\boldsymbol{D}\) as stiffness and damping matrices (Table~\ref{tab:controller-params}). 
The damping matrix \(\boldsymbol{D}\) is computed using \(\bf \sqrt{\Lambda}\sqrt{K} + \sqrt{K}\sqrt{\Lambda}\).

Finally, the optimal joint accelerations yield the torques
\begin{gather}
\boldsymbol{\tau} = \boldsymbol{M}(\boldsymbol{\mathrm{q}})\,\boldsymbol{\mathrm{\ddot q}}^* + \boldsymbol h(\boldsymbol{\mathrm{ q}},\boldsymbol{\mathrm{\dot q}}),
\end{gather}
which are applied to the simulator.  
This converts the MPC generated tray tilts into dual–arm torques realizing the desired motion while respecting all constraints.

\section{EXPERIMENTS}
\label{sec:experiments}
We evaluate our proposed framework as a simulated dual-arm non-prehensile tray manipulation task, 
where two UFactory xArm7 manipulators jointly tilt a tray to drive an object on the tray towards a desired target state.
We emphasize robustness to variations in object mass, geometry, and tray–object friction. 
All experiments are conducted in the MuJoCo simulator with identical physical models, initial conditions, and prediction horizons to ensure unbiased comparisons across the methods.  

\subsection{Task Setup}
\label{sec:task setup}

In each trial, a tray with mass of 1\,kg and dimensions of \(40\,\text{cm} \times 30\,\text{cm} \times 1\,\text{cm}\), is manipulated by the control input \(\boldsymbol{\mathrm{u}}\) executed as in Sec.~\ref{sec:davil-controller}.

We consider three object geometries: \{\text{cube}, \text{cylinder}, \text{sphere}\}, 
representing area, line, and point contacts, respectively. Each object is evaluated at two different masses \(\{1,2\}\,\text{kg}\) 
and three different friction coefficients \(\{0.05,0.10,0.20\}\), 
resulting in \(3 \times 3 \times 2 = 18\) unique object configurations.
Fixing the initial object position to be at the tray's center, we uniformly sample target positions at a radial distance of $\left[0.08\ ,0.12 \right]$ m from the center and average over them to get the final metrics for each configuration.
The xArm7 bases are mounted with a fixed separation of \(1.4\ \text{m}\), 
providing sufficient dexterous workspace for coordinated tray manipulation.

\begin{table}[h]
\vspace{6pt}
\centering
\captionsetup{font=footnotesize}
\begin{tabular}{lll}
\hline
\textbf{Symbol} & \textbf{Description} & \textbf{Value} \\
\hline
\(T_{s}\) & Simulation/MPC time step & 0.002s \\
\(\mathrm{n}_\mathrm{x}\) & State dimension & 8 \\
\(\mathrm{n}_\mathrm{u}\) & Control dimension & 2 \\
\(\mathrm{N}\) & Prediction horizon & 20 \\
\(\boldsymbol{\mathrm{u}}_{\alpha}^{\max},\boldsymbol{\mathrm{u}}_{\beta}^{\max}\) & Max. roll/pitch command & \(0.6\,\text{rad}\) \\
\(\boldsymbol{\mathrm{Q}_{\mathrm{p}}}\) & Pose error weight & 250 $\boldsymbol{\mathrm{I}_{6}}$\\
\(\boldsymbol{\mathrm{Q}_{\mathrm{v}}}\) & Velocity error weight & $2\ \boldsymbol{\mathrm{I}_{6}}$\\
\(\boldsymbol{\mathrm{Q}_{\mathrm{R}}}\) & Control effort weight & $0.2\ \boldsymbol{\mathrm{I}_{4}}$ \\
\(\boldsymbol{K}\) & Cartesian stiffness (N/m) & \(\mathrm{diag}(5000\  \mathbf{I}_3, 50\ \mathbf{I}_3)\)\\
\(\boldsymbol{K}_{\mathrm{null}}\) & Joint stiffness (posture) & \(7\  \boldsymbol{I}_3\,\text{N/m}\) \\
\hline
\end{tabular}
\caption{\small Controller parameters used in all experiments.}
\label{tab:controller-params}
\end{table}

\subsection{Controller Parameters}

The weight matrices used by the MPC in the high-level task, as well as those used by the low-level QP based impedance controller are listed in Table~\ref{tab:controller-params}. These are the default values used in all experiments in Sec.~\ref{sec:experiments} unless mentioned otherwise.



The weight matrices $\boldsymbol{\mathrm{Q}_{\mathrm{p}}}$, $\boldsymbol{\mathrm{Q}_{\mathrm{v}}}$, $\boldsymbol{\mathrm{Q}_{\mathrm{N}}}$, and $\boldsymbol{\mathrm{Q}_{\mathrm{R}}}$ are tuned iteratively. 
Increasing $\boldsymbol{\mathrm{Q}_{\mathrm{p}}}$ places greater emphasis on accurate final positioning of the object, while higher $\boldsymbol{\mathrm{Q}_{\mathrm{v}}}$ helps dampen overshoot. 
$\boldsymbol{\mathrm{Q}_{\mathrm{N}}}$ enforces convergence to the desired state, and $\boldsymbol{\mathrm{Q}_{\mathrm{R}}}$ penalizes excessive torque changes, promoting smooth and feasible tray motions without saturating the torques.This tuning balances positioning accuracy with control effort. 


\subsection{Training Details for LMPC}
\label{sec:training}
The reinforcement learning policy was trained in simulation using MuJoCo, running three environments in parallel and collecting 500 episodes of 20,000 steps each---sufficient to observe convergence given an initial, arbitrary, non-zero parameter vector. We used PPO with a learning rate of $3 \times 10^{-4}$, value function coefficient $c_v = 0.5$, and entropy coefficient $\beta = 0.01$. Training covered a total of $500 \times 20{,}000$ steps ($\approx 2$ hours of wall-clock time) and was evaluated every $2048$ iterations. 
Further, we have adopted domain randomization to randomly select any one of the 18 physical object configurations to improve the generalization capabilities of the RL agent.

\begin{table*}[ht]
\vspace{6pt}
\centering
\scriptsize
\captionsetup{font=footnotesize}
\resizebox{\textwidth}{!}{%
\begin{tabular}{|c|c|c|ccc|ccc|ccc|}
\hline
\multirow{3}{*}{\textbf{Object}} & \multirow{3}{*}{\textbf{Mass(Kg)}} & \multirow{3}{*}{\textbf{Friction}} 
& \multicolumn{3}{c|}{\textbf{Settling Time(s)}} 
& \multicolumn{3}{c|}{\textbf{Steady State Error(m)}} 
& \multicolumn{3}{c|}{\textbf{Control Effort}} \\
\cline{4-12}
 &  &  
 & \textbf{PMPC} & \textbf{RMPC} & \textbf{LMPC} 
 & \textbf{PMPC} & \textbf{RMPC} & \textbf{LMPC} 
 & \textbf{PMPC} & \textbf{RMPC} & \textbf{LMPC} \\
\hline

\multirow{6}{*}{Cube}
  & \multirow{3}{*}{1}
    & 0.05 & 6.70 & 17.14 & \textbf{3.89}
    & 0.060 & 0.017 & \textbf{0.0066}
    & 2174.35 & \textbf{943.46} & 2007.70 \\
  &  & 0.1  & 7.34 & 16.92 & \textbf{3.47}
    & 0.020 & 0.018 & \textbf{0.0075}
    & 2131.62 & \textbf{949.57} & 1885.27 \\
  &  & 0.2  & 7.43 & 17.02 & \textbf{2.57}
    & 0.020 & 0.026 & \textbf{0.016}
    & 2209.89 & \textbf{1164.74} & 1884.86 \\
\cline{2-12}
  & \multirow{3}{*}{2}
    & 0.05 & 7.16 & 21.11 & \textbf{4.85}
    & 0.050 & 0.017 & \textbf{0.0059}
    & 2328.56 & 3873.42 & \textbf{2085.31} \\
  &  & 0.1  & 6.52 & 21.70 & \textbf{4.91}
    & 0.039 & 0.013 & \textbf{0.0055}
    & \textbf{2192.75} & 3906.57 & 2258.05 \\
  &  & 0.2  & 6.89 & 21.30 & \textbf{3.86}
    & 0.039 & 0.014 & \textbf{0.0117}
    & 2275.39 & 3941.35 & \textbf{2182.49} \\
\hline

\multirow{6}{*}{Cylinder}
  & \multirow{3}{*}{1}
    & 0.05 & 6.24 & 16.53 & \textbf{4.02}
    & 0.060 & 0.025 & \textbf{0.0030}
    & 1704.56 & \textbf{992.04} & 1780.91 \\
  &  & 0.1  & 5.92 & 16.60 & \textbf{4.22}
    & 0.031 & 0.021 & \textbf{0.0026}
    & 1718.19 & \textbf{893.93} & 1782.94 \\
  &  & 0.2  & 4.97 & 18.08 & \textbf{4.74}
    & \textbf{0.0019} & 0.020 & 0.0044
    & 1172.72 & \textbf{924.44} & 2041.98 \\
\cline{2-12}
  & \multirow{3}{*}{2}
    & 0.05 & 4.66 & 15.82 & \textbf{3.22}
    & 0.0315 & 0.013 & \textbf{0.0044}
    & \textbf{1569.18} & 2925.26 & 2063.84 \\
  &  & 0.1  & 4.12 & 19.48 & \textbf{2.93}
    & \textbf{0.0024} & 0.014 & 0.0044
    & \textbf{1582.98} & 2855.69 & 1988.34 \\
  &  & 0.2  & 4.96 & 18.61 & \textbf{4.12}
    & 0.040 & 0.012 & \textbf{0.0041}
    & \textbf{1659.26} & 2970.69 & 1926.88 \\
\hline

\multirow{6}{*}{Sphere}
  & \multirow{3}{*}{1}
    & 0.05 & 5.39 & 15.58 & \textbf{1.48}
    & \textbf{0.0052} & 0.034 & 0.0166
    & 1817.05 & \textbf{1102.22} & 2007.70 \\
  &  & 0.1  & 4.27 & 13.74 & \textbf{1.21}
    & \textbf{0.0047} & 0.030 & 0.0207
    & 1651.87 & \textbf{998.74} & 1885.27 \\
  &  & 0.2  & 4.30 & 12.45 & \textbf{1.39}
    & \textbf{0.0060} & 0.030 & 0.0223
    & 1735.35 & \textbf{1164.49} & 1884.86 \\
\cline{2-12}
  & \multirow{3}{*}{2}
    & 0.05 & 6.05 & 11.97 & \textbf{1.34}
    & 0.054 & 0.028 & \textbf{0.0304}
    & \textbf{2068.16} & 3461.95 & 2085.31 \\
  &  & 0.1  & 4.65 & 10.83 & \textbf{1.20}
    & \textbf{0.0054} & 0.031 & 0.0301
    & \textbf{2041.12} & 3527.24 & 2258.05 \\
  &  & 0.2  & 5.76 & 12.80 & \textbf{0.82}
    & \textbf{0.0137} & 0.027 & 0.0262
    & \textbf{2107.73} & 3515.69 & 2182.49 \\
\hline
\end{tabular}}
\caption{\small Performance comparison of different object geometries with varying mass and friction parameters, each configuration averaged over 20 radially sampled targets.}
\label{tab:performance_metrics}
\end{table*}

\subsection{Implementation Details}
\label{sec:implementation_details}
Both RMPC and LMPC are implemented in Python using the PyTorch framework. The nonlinear MPC and the optimization-based dual arm controller are solved using the Interior Point Optimization (IPOPT) solver in CasADi~\cite{casadi}. All simulations and training experiments were executed on a Ryzen 9 8945HS CPU and NVIDIA GeFORCE RTX 4090 GPU.

To ensure real–time performance during training and evaluation, the codebase was parallelized using Python’s \texttt{multiprocessing} library.

Specifically, the CasADi solver for the MPC and the dual–arm controller and the MuJoCo simulation environment were each executed in separate processes, which communicate via a shared memory buffer in CPU memory. Synchronization is achieved using multiple mutex locks, with the MuJoCo solver's updates as a reference clock.


\subsection{Evaluation Metrics}
\label{sec:Evaluation Metrics}
The data collected for evaluation is sampled at a frequency of 500Hz for a duration of 20 seconds. At each timestep $i \in [1,\mathrm{n}]$, the position error is defined as $\boldsymbol{\mathrm{e}}_i = \mathbf{p}^{\mathrm{ref}} - \mathbf{p}_i$, with $\mathrm{n}$ the total number of steps.
We assess the three MPC methods (PMPC, RMPC, LMPC) using three widely used performance measures~\cite{ogata2010modern,skogestad2005multivariable}.

\begin{itemize}
  \item \textbf{Steady State Error:} Measures the mean error over the last 4s (20 \% of the samples) as all three methods reach steady state before this. It is computed as:
  \begin{equation}
  \hat{\mathrm{e}}_{\text{ss}} = \frac{1}{\text{0.2n}}\sum_{\text{$i$=0.8n+1}}^\text{n}\|\text{e}_\text{$i$}\|^2
  \end{equation}
   

  \item \textbf{Settling Time:}  
  Measures the time taken to reach the steady state with a tolerance of 1\%. It is defined as the smallest time index $\text{$i$}_s$ such that
  \begin{equation}
  \lVert \text{e}_\text{$i$}\rVert \le \varepsilon, \quad \forall\,\text{$i$}\ge \text{$i$}_s
  \end{equation} 
  where $\varepsilon=1.01\lVert \hat{\mathrm{e}}_{\text{ss}}\rVert$.

  \item \textbf{Control Effort:}  
  Evaluates the efficiency and smoothness of the applied torques.  
  We compute energy and rate costs from the joint torques $\boldsymbol{\tau}_\text{k}$ of both xArm7s as 
  \begin{equation}
  \mathrm{J_{\text{effort}}} = \text{w}_\text{E}\sum_{\text{$i$}=1}^{n}\|\boldsymbol{\tau}_\text{$i$}\|^{2}\,\Delta t +\text{w}_\text{R}\sum_{\text{$i$}=2}^{\text{n}}\|\frac{\boldsymbol{\tau}_\text{$i$}-\boldsymbol{\tau}_\text{$i$-1}}{\Delta t}\|^{2},
  \end{equation} 

where $\text{w}_\text{E}$, $\text{w}_\text{R}$ are the associated weights. 

\end{itemize}

\section{RESULTS}


We evaluate the performance of \acro{} across a range of object configurations using the metrics defined in Sec.~\ref{sec:Evaluation Metrics}.

\begin{figure}[h]
    \centering
    \captionsetup{font=footnotesize}
    \includegraphics[width=\linewidth]{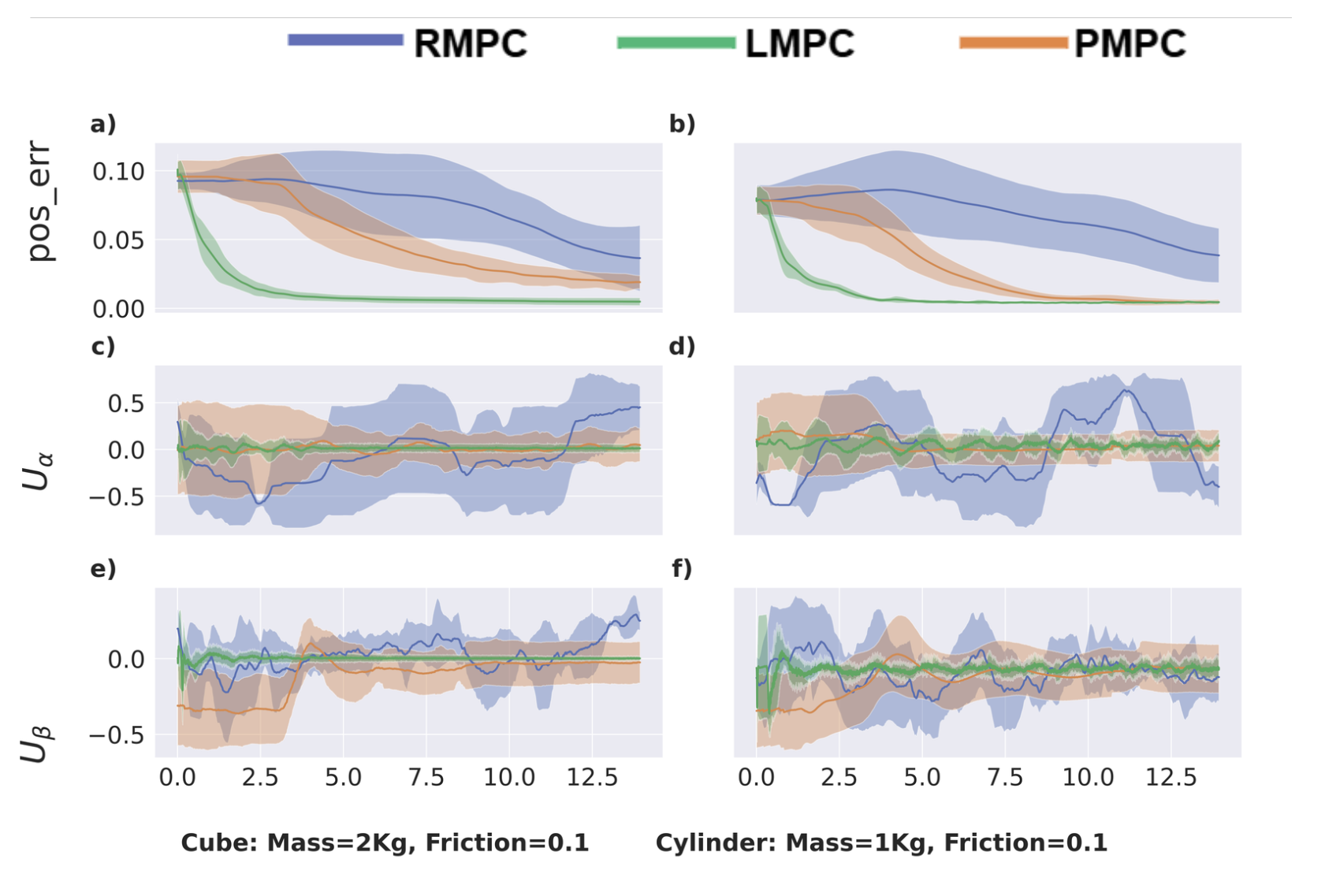}
    \caption{Performance of  two objects, Cube on the left and Cylinder on the right  visualized as a standard deviation and average plot denoting the position error and tilt commands given by the LMPC RMPC and PMPC. The color legend corresponding to the three methods shown on top }
    \label{fig:quantitative}
\end{figure}

\subsection{Quantitative Results}
Table~\ref{tab:performance_metrics} summarizes the performance of all three controllers across all the object configurations described in Sec.~\ref{sec:task setup}.
We observe that LMPC has better settling time and  and steady state error. However, it's control effort is relatively high. RMPC requires low control effort for low mass objects but it shows higher settling time. PMPC offers better steady state error for rolling contacts and low control effort for high mass rolling contacts. 

We further observe that RMPC and LMPC generalize well across objects, masses, and friction coefficients, which can be observed in their consistent settling times and steady state errors. PMPC cannot generalize and requires object-specific parameters  to be provided. This is observed in Table~\ref{tab:performance_metrics}.

While LMPC outperforms the other two methods in terms of settling times, it requires training and computation~\ref{sec:training}. RMPC adapts real time and does not require any training.

Fig. \ref{fig:quantitative} visualizes the above observations for 2 different object configurations, showing the trends in steady state error across RMPC, LMPC and PMPC, and the respective commands produced by these methods. This portrays the trends observed across Table \ref{tab:performance_metrics}. 
LMPC converges the fastest while PMPC shows a smooth settling to the lowest steady state error for the Cube. 

\begin{figure}[h]
    \centering
    \captionsetup{font=footnotesize}
    \includegraphics[width=0.75\linewidth]{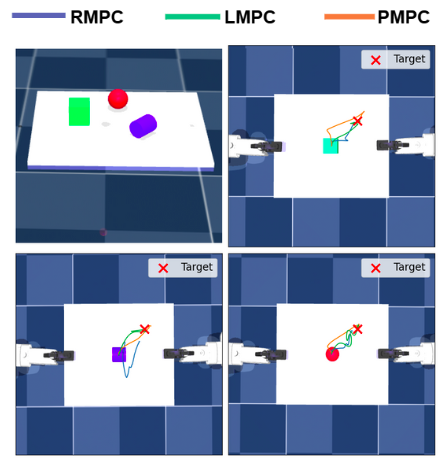}
    
    
    \caption{\small \textbf{Visualizations of the trajectories:} This figure depicts the trajectories followed by a cube(green), sphere(red), and a cylinder(purple) with mass 1kg and friction 0.2 , manipulated by using 3 different MPC methods.
    The curves shows only the path of the object taken on the tray, longer path doesn't mean longer convergence. The color legend corresponds to the three methods as shown on top.}

    \label{fig:qualitative}
\end{figure}

\subsection{Qualitative Results}We qualitatively evaluate our framework across diverse and challenging scenarios. Fig.~\ref {fig:qualitative} shows trajectories for varying object shapes, each presenting a different manipulation challenge due to different contact interactions. We chose a shorter goal to observe the trajectories given by different controllers to test the ability to react faster. The controller behaviours are clearly differentiated. PMPC takes long (Fig.~\ref {fig:qualitative}) yet faster (Table~\ref {tab:performance_metrics}) trajectories which can be treated a trade-off of simplified contact dynamics model as Eq. \ref{eq:object_dynamics}. RMPC converges towards the target while adapting online; however, in some scenarios, it overshoots with a sub-optimal estimate of dynamics but manages to converge as it updates it parameters. As observed in the Fig. \ref{fig:qualitative} cylinder shows some steady state error of 2-2.3 cm which is consistent with Table \ref{tab:performance_metrics}. LMPC is always taking the shortest and fastest trajectories compared to the other two methods(Fig.~\ref {fig:qualitative}) with good overall performance along with genaralization over different object configurations.
These rollouts highlight how environment parameters strongly affect manipulation dynamics, underscoring the need for controllers that generalize across conditions rather than overfit to a single setup.
\section{CONCLUSION AND FUTURE WORK}
This paper presents \acro{}, a first framework for dual-arm non-prehensile tray manipulation. By integrating nonlinear MPC with optimization-based impedance control, our approach enables object transportation with accuracy up to \(2\ \text{mm}\) on dynamically controlled trays, a capability critical for service robots. Our comparison of three dynamics modeling strategies reveal various strengths: physics-based models offer interpretability, regression-based models provide online adaptation, while reinforcement learning delivers superior generalization with faster convergence.

Through validation across objects with varying mass, geometry, and friction properties, we demonstrate that \acro{} effectively navigates the complex, coupled dynamics between dual manipulators and sliding objects. Future work will extend this framework to handle surfaces with variable friction, uneven surfaces, complex objects, orientation control through active yawing, and non-planar object manipulation.



\bibliographystyle{unsrt}
\bibliography{bibTEX}

\addtolength{\textheight}{-12cm}   





\end{document}